\documentclass{article}

\usepackage{microtype}
\usepackage{graphicx}
\usepackage{subfigure}
\usepackage{booktabs} % for professional tables
\usepackage{tabularx}

\usepackage{hyperref}

\usepackage[accepted]{icml2019}

\icmltitlerunning{MMLSpark: Unifying Machine Learning Ecosystems at Massive Scales}

\begin{document}

\twocolumn[
\icmltitle{MMLSpark: Unifying Machine Learning Ecosystems at Massive Scales}

% It is OKAY to include author information, even for blind
% submissions: the style file will automatically remove it for you
% unless you've provided the [accepted] option to the icml2019
% package.

% List of affiliations: The first argument should be a (short)
% identifier you will use later to specify author affiliations
% Academic affiliations should list Department, University, City, Region, Country
% Industry affiliations should list Company, City, Region, Country

% You can specify symbols, otherwise they are numbered in order.
% Ideally, you should not use this facility. Affiliations will be numbered
% in order of appearance and this is the preferred way.
\icmlsetsymbol{equal}{*}

\begin{icmlauthorlist}
  \icmlauthor{Mark Hamilton}{aai}
  \icmlauthor{Sudarshan Raghunathan}{aair}
  \icmlauthor{Ilya Matiach}{aml}
  \icmlauthor{Andrew Schonhoffer}{aml}
  \icmlauthor{Anand Raman}{aair}
  \icmlauthor{Eli Barzilay}{aai}
  \icmlauthor{Karthik Rajendran}{maidap, equal}
  \icmlauthor{Dalitso Banda}{maidap, equal}
  \icmlauthor{Casey Jisoo Hong}{maidap, equal}
  \icmlauthor{Manon Knoertzer}{maidap, equal}
  \icmlauthor{Ben Brodsky}{aair}
  \icmlauthor{Minsoo Thigpen}{maidap}
  \icmlauthor{Janhavi Suresh Mahajan}{maidap}
  \icmlauthor{Courtney Cochrane}{maidap}
  \icmlauthor{Abhiram Eswaran}{maidap}
  \icmlauthor{Ari Green}{maidap}
\end{icmlauthorlist}

\icmlaffiliation{aai}{Microsoft Applied AI, Cambridge, Massachusetts, USA}
\icmlaffiliation{aair}{Microsoft Applied AI, Redmond, Washington, USA}
\icmlaffiliation{maidap}{Microsoft AI Development Acceleration Program, Cambridge, Massachusetts, USA}
\icmlaffiliation{aml}{Microsoft Azure Machine Learning,  Cambridge, Massachusetts, USA}

\icmlcorrespondingauthor{Mark Hamilton}{marhamil@microsoft.com}

% You may provide any keywords that you
% find helpful for describing your paper; these are used to populate
% the "keywords" metadata in the PDF but will not be shown in the document
\icmlkeywords{Apache Spark, Deep Learning, Microservices, Container Orchestration, Kubernetes, Machine Learning Frameworks, CNTK, LightGBM, Distributed Computing, SQL, Streaming, Web Serving, Wildlife Conservation, Model Explainability, LIME}

\vskip 0.3in
]

% this must go after the closing bracket ] following \twocolumn[ ...

% This command actually creates the footnote in the first column
% listing the affiliations and the copyright notice.
% The command takes one argument, which is text to display at the start of the footnote.
% The \icmlEqualContribution command is standard text for equal contribution.
% Remove it (just {}) if you do not need this facility.

%\printAffiliationsAndNotice{}  % leave blank if no need to mention equal contribution
\printAffiliationsAndNotice{\icmlEqualContribution} % otherwise use the standard text.

\begin{abstract}
We introduce Microsoft Machine Learning for Apache Spark (MMLSpark), an open-source library that expands the Apache Spark distributed computing library to tackle problems in deep learning, micro-service orchestration, gradient boosting, model interpretability, and other areas of modern machine learning. We also present a novel machine learning deployment system called Spark Serving that can deploy Apache Spark programs as distributed, sub-millisecond latency web services with significantly greater flexibility and lower latencies than existing frameworks. Spark Serving generalizes beyond just map-style computations and allows distributed aggregations, joins, and shuffles and allows users to leverage the same cluster for both training and deployment. Our contributions allow easy composition across machine learning frameworks, compute modes (batch, streaming, and RESTful web serving) and cluster types (static, elastic, and serverless). We demonstrate the value of MMLSpark by creating a method for deep object detection capable of learning without human labeled data and demonstrate its effectiveness for Snow Leopard conservation. We also demonstrate its ability to create large-scale image search engines.
\end{abstract}

\section{Introduction}
\label{Introduction}
As the field of machine learning has advanced, frameworks for using, authoring, and training machine learning systems have proliferated. These different frameworks often have dramatically different APIs, data models, usage patterns, and scalability considerations. This heterogeneity makes it difficult to combine systems and complicates production deployments. In this work, we present Microsoft Machine Learning for Apache Spark (MMLSpark), an ecosystem that aims to unify major machine learning workloads into a single API for execution in a variety of distributed production grade environments and languages. We describe the techniques and principles used to unify a representative sample of machine learning technologies, each with its own software stack, communication requirements, and paradigms. We also introduce tools for deploying these technologies as distributed real-time web services. Code and documentation for MMLSpark can be found through our website, \url{https://aka.ms/spark}.

\section{Background}
\label{Background}

Throughout this work we build upon the distributed computing framework Apache Spark \cite{spark}. Spark is capable of a broad range of workloads and applications such as fault-tolerant and distributed map, reduce, filter, and aggregation style programs. Spark improves on its predecessors MapReduce and Hadoop by reducing disk IO with in memory computing, and whole program optimization \cite{mapreduce, hadoop}. Spark clusters can adaptively resize to compute a workload efficiently (elasticity) and can run on resource managers such as Yarn, Mesos, Kubernetes, or manually created clusters. Furthermore, Spark has language bindings in several popular languages like Scala, Java, Python, R, Julia, C\# and F\#, making it usable from almost any project.   

In recent years, Spark has expanded its scope to support SQL, streaming, machine learning, and graph style computations \cite{sparksql, sparkml, graphx}. This broad set of APIs allows a rich space of computations that we can leverage for our work. More specifically, we build upon the SparkML API, which is similar to the popular Python machine learning library, scikit-learn \cite{sklearn}. Like scikit-learn, all SparkML models have the same API, which makes it easy to create, substitute, and compose machine learning algorithms into ``pipelines''. However, SparkML has several key advantages such as limitless scalability, streaming compatibility, support for structured datasets, broad language support, a fluent API for initializing complex algorithms, and a type system that differentiates computations based on whether they extract state (learn) from data. In addition, Spark clusters can use a wide variety of hardware SKUs making it possible to leverage modern advances in GPU accelerated frameworks like Tensorflow, CNTK, and PyTorch \cite{tensorflow, cntk, pytorch}. These properties make the SparkML API a natural and principled choice to unify the APIs of other machine learning frameworks.

Across the broader computing literature, many have turned to intermediate languages to ``unify'' and integrate disparate forms of computation. One of the most popular of these languages is the Hypertext Transfer Protocol (HTTP) used widely throughout internet communications. To enable broad adoption and integration of code, one simply needs to create a web-hosted HTTP endpoint or ``service''. Putting compute behind an intermediate language allows different system components to scale independently to minimize bottlenecks. If services reside on the same machine, one can use local networking capabilities to bypass internet data transfer costs and come closer to the latency of normal function dispatch. This pattern is referred to as a ``micro-service'' architecture, and powers many of today's large-scale applications \cite{microservice1}.

Many machine learning workflows rely on deploying learned models as web endpoints for use in front-end applications. In the Spark ecosystem, there are several ways to deploy applications as web services such as Azure Machine Learning Services (AML), Clipper, and MLeap. However, these frameworks all compromise on the breadth of models they export, or the latency of their deployed services. AML deploys PySpark code in a dockerized Flask application that uses Spark's Batch API on a single node standalone cluster \cite{azure-machine-learning, flask}. Clipper uses on an intermediate RPC communication service to invoke a Spark batch job for each request \cite{clipper}. Both method use Spark's Batch API which adds large overheads. Furthermore, if back-end containers are isolated, this precludes services with inter-node communication like shuffles and joins. MLeap achieves millisecond latencies by re-implementing SparkML models in single threaded Scala, and exporting SparkML pipelines to this alternate implementation \cite{mleap}. This incurs a twofold development cost, a lag behind the SparkML library, and a export limitation to models in Spark's core, which is a small subset of the ecosystem. In section \ref{serving}, we present a novel method, Spark Serving, that achieves millisecond latencies without compromising the breadth or latency of models and does not rely on model export. This makes the transition from distributed training to distributed serving seamless and instant.

Many companies such as Microsoft, Amazon, IBM, and Google have embraced model deployment with web services to provide pre-built intelligent algorithms for a wide range of applications \cite{aws-service, msft-service, ibm-watson}. This standardization enables easy use of cloud intelligence and abstracts away implementation details, environment setup, and compute requirements. Furthermore, intelligent services allow application developers to quickly use existing state of the art models to prototype ideas. In the Azure ecosystem, the Cognitive Services provide intelligent services in domains such as text, vision, speech, search, time series, and geospatial workloads.

\section{Contributions}
\label{Contributions}

In this section we describe our contributions in three key areas: 1) Unifying several Machine Learning ecosystems with Spark. 2) Integrating Spark with the networking protocol HTTP and several intelligent web services. 3) Deploying Spark computations as distributed web services with Spark Serving.

These contributions allow users to create scalable machine learning systems that draw from a wide variety of libraries and expose these contributions as web services for others to use. All of these contributions carry the common theme of building a single distributed API that can easily and elegantly create a variety of different intelligent applications. In Section \ref{Applications} we show how to combine these contributions to solve problems in unsupervised object detection, wildlife ecology, and visual search engine creation.

\subsection{Algorithms and Frameworks Unified in MMLSpark}
\label{Contributions:AandF}

\subsubsection{Deep Learning}

To enable GPU accelerated deep learning on Spark, we have previously parallelized Microsoft's deep learning framework, the Cognitive Toolkit (CNTK) \cite{cntk, mmlspark}. This framework powers roughly 80\% of Microsoft's internal deep learning workloads and is flexible enough to create most models described in the deep learning literature. CNTK is similar to other automatic differentiation systems like Tensorflow, PyTorch, and MxNet as they all create symbolic computation graphs that automatically differentiate and compile to machine code. These tools liberate developers and researchers from the difficult task of deriving training algorithms and writing GPU accelerated code.

CNTK's core functionality is written in C++ but exposed to C\# and Python through bindings. To integrate this framework into Spark, we used the Simple Wrapper and Interface Generator (SWIG) to contribute a set of Java Bindings to CNTK \cite{swig}. These bindings allow users to call and train CNTK models from Java, Scala and other JVM based languages. We used these bindings to create a SparkML transformer to distribute CNTK in Scala. Additionally we automatically generate PySpark and SparklyR bindings for all of MMLSpark's Scala transformers, so all contributions in this work are usable across different languages. To improve our implementation's performance we broadcast the model to each worker using Bit-Torrent broadcasting, re-use C++ objects to reduce garbage collection overhead, asynchronously mini-batch data, and share weights between local threads to reduce memory overhead. With CNTK on Spark, users can embed any deep network into parallel maps, SQL queries, and streaming pipelines. We also contribute and host a large cloud repository of trained models and tools to perform image classification with transfer learning. We have utilized this work for wildlife recognition, bio-medical entity extraction, and gas station fire detection \cite{mmlspark}.

We released our implementation concurrently with Databrick's ``Deep Learning Pipelines'' that provides an analogous integration of Spark and Tensorflow \cite{deep-learning-pipelines}. Our two integrations share the same API making it easy to use CNTK and/or Tensorflow inside of SparkML pipelines without code changes. 

\subsubsection{Gradient Boosting and Decision Trees}

Though Tensorflow and CNTK provide GPU enabled deep networks, these frameworks are optimized for differentiable models. To efficiently learn tree/forest-based models, many turn to GPU enabled gradient boosting libraries such as XGBoost or LightGBM \cite{xgboost, lightgbm}. We contribute an integration of LightGBM into Spark to enable large scale optimized gradient boosting within SparkML pipelines. LightGBM is one of the most performant decision tree frameworks and can use socket or Message Passing Interface (MPI) communication schemes that communicate much more efficiently than SparkML's Gradient Boosted Tree implementation. As a result, LightGBM trains up to 30\% faster than SparkML's gradient boosted tree implementation and exposes many more features, optimization metrics, and growing/pruning parameters. Like CNTK, LightGBM is written in C++ and publishes bindings for use in other languages. For this work, we used SWIG to contribute a set of Java bindings to LightGBM for use in Spark. Unlike our work with CNTK on Spark, LightGBM training involves nontrivial MPI communication between workers. To unify Spark's API with LightGBM's communication scheme, we transfer control to LightGBM with a Spark ``MapPartitions'' operation. More specifically, we communicate the hostnames of all workers to the driver node of the Spark cluster and use this information to launch an MPI ring. The Spark worker processes use the Java Native Interface to transfer control and data to the LightGBM processes. This integration allows users to create performant models for classification, quantile regression, and other applications that excel in discrete feature domains.

\subsubsection{Model Interpretability}

In addition to integrating frameworks into Spark through transfers of control, we have also expanded SparkML's native library of algorithms. One example is our distributed implementation of Local Interpretable Model Agnostic Explanations (LIME) \cite{lime}. LIME provides a way to ``interpret'' the predictions of $\bf{any}$ model without reference to that model's functional form. More concretely, LIME interprets black box functions though a locally linear approximation constructed from a sampling procedure. LIME, and its generalization SHAP, rely solely on function evaluations and can and apply to any black box algorithm \cite{shap}. We consider other methods such as DeepLIFT and feature gradients outside the scope of this work because SparkML models are not necessarily differentiable \cite{deep-lift}.

Intuitively speaking, if ``turning off'' a part of the model input dramatically changes a model's output, that part is ``important''. More concretely, LIME for image classifiers creates thousands of perturbed images by setting random chunks or ``superpixels'' of the image to a neutral color. Next, it feeds each of these perturbed images through the model to see how the perturbations affect the model's output. Finally, it uses a locally weighted Lasso model to learn a linear mapping between a Boolean vector representing the ``states'' of the superpixels to the model's outputs. Text and tabular LIME employ similar featurization schemes, and we refer readers to \cite{lime} for detailed descriptions. 

To interpret a classifier's predictions for an image, one must evaluate the classifier on thousands of perturbed images to sufficiently sample the superpixel state space. Practically speaking, if it takes 1 hour to score a model on your dataset, it would take $\approx50$ days to interpret this dataset with LIME. We have created a distributed implementation to reduce this massive computation time. LIME affords several possible distributed implementations, and we have chosen a parallelization scheme that speeds each individual interpretation. More specifically, we parallelize the superpixel decompositions over the input images. Next, we iterate through the superpixel decompositions and create a new parallel collection of ``state samples'' for each input image. We then perturb these images and apply the model in parallel. Finally, we fit a distributed linear model to the inner collection and add its weights to the original parallel collection. Because of this nontrivial parallelization scheme, this kind of integration benefited from a complete re-write in fast compiled Scala and Spark SQL, as opposed to using a tool like Py4J to integrate the existing LIME repository into Spark. 
\subsection{Unifying Microservices with Spark}

In Section \ref{Contributions:AandF} we explored three contributions that unify Spark with other Machine Learning tools using the Java Native Interface (JNI) and function dispatch. These methods are efficient, but require re-implementing code in Scala or auto-generating wrappers from existing code. For many frameworks, these dispatch-based integrations are impossible due to differences in language, operating system, or computational architecture. For these cases, we can utilize inter-process communication protocols like HTTP to bridge the gap between systems. 

\subsubsection{HTTP on Spark}
We present HTTP on Spark, an integration between the entire HTTP communication protocol and Spark SQL. HTTP on Spark allows Spark users to leverage the parallel networking capabilities of their cluster to integrate any local, docker, or web service. At a high level, HTTP on Spark provides a simple and principled way to integrate $\bf{any}$ framework into the Spark ecosystem. The contribution adds HTTP Request and Response types to the Spark SQL schema so that users can create and manipulate their requests and responses using SQL operations, maps, reduces, and filters. When combined with SparkML, users can chain services together, allowing Spark to function as a distributed micro-service orchestrator. HTTP on Spark also automatically provides asynchronous parallelism, batching, throttling, and exponential back-offs for failed requests. 

\subsubsection{The Cognitive Services on Spark}

We have built on HTTP on Spark to create a simple and powerful integration between the Microsoft Cognitive Services and Spark. The Cognitive Services on Spark allows users to embed general purpose and continuously improving intelligent models directly into their Spark and SQL computations. This contribution aims to liberate users from low level networking details, so they can focus on creating intelligent distributed applications. Each Cognitive Service is a SparkML transformer, so users can add services to existing SparkML pipelines. We introduce a new class of model parameters to the SparkML framework that allow users to parameterize models by either a single scalar value or vectorize the requests with columns of the DataFrame. This syntax yields a succinct yet powerful fluent query language that offers a full distributed parameterization without clutter. For example, by vectorizing the ``subscription key'' parameter, users can distribute requests across several accounts, regions, or deployments to maximize throughput and resiliency to error.

\begin{figure}[ht]
\vskip 0.1in
\begin{center}
\centerline{
\includegraphics[width=3in]{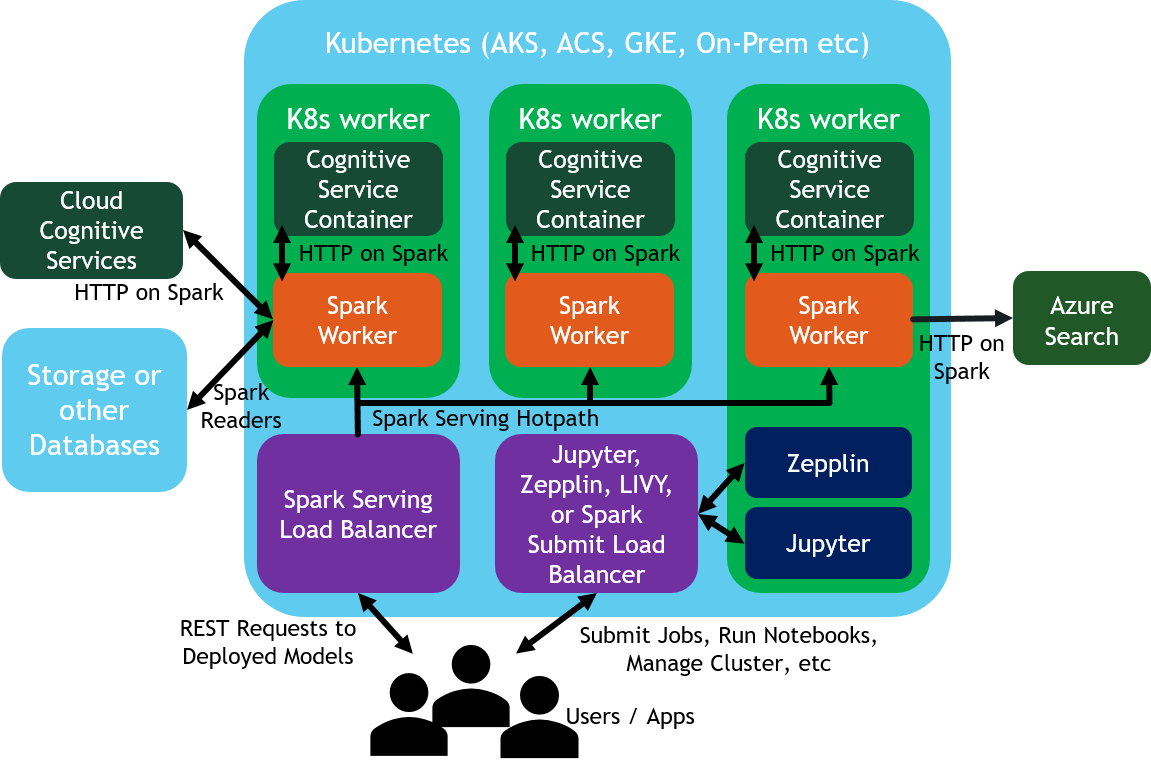}
}
\caption{Architecture diagram of our integration of cloud and containerized Cognitive Services. Architecture depicted on Kubernetes, but any container orchestrator could deploy the same. Load balancers expose deployed Spark Serving models, job submission endpoints, and monitoring frontends. This helm chart can also leverage our integration between Azure Search and Spark. Note that we omit Kubernetes and Spark head nodes for simplicity.}
\label{fig:k8s-architecture}
\end{center}
\vskip -0.4in
\end{figure}

Once can combine HTTP on Spark with Kubernetes or other container orchestrators to deploy services directly onto Spark worker machines \cite{Kubernetes}. This enables near native integration speeds as requests do not have to travel across machines. The cognitive services on Spark can also call the newly released containerized cognitive services, which dramatically reduces the latency of cognitive service pipelines. We have contributed a helm chart for deploying a Spark based microservice architecture with containerized cognitive services, load balancers for Spark Serving, and integration with the newly released second generation of Azure Storage with a single command \cite{azure-storage-gen2, helm}. Figure \ref{fig:k8s-architecture} shows a diagram of the aforementioned architecture. 

\subsubsection{Azure Search Sink for Spark}

We demonstrate the flexibility and robustness of the HTTP on Spark framework by contributing an integration between Spark and Azure Search. Azure Search is a cloud database that supports rapid information retrieval and query execution on heterogeneous, unstructured data \cite{azure-search}. Azure Search leverages elastic search to index documents and provide REST APIs for document search on linguistic similarity and a variety of other filters and logical constraints. With this integration, users can leverage the frameworks mentioned in \ref{Contributions:AandF} to enrich their documents in parallel prior to indexing. 

\subsection{Spark Serving: Scalable Real-Time Web Serving}
\label{serving}

Through HTTP on Spark, we have enabled Spark as a distributed web client. In this work we also contribute Spark Serving, a framework that allows Spark clusters to operate as distributed web $\it{servers}$. Spark Serving builds upon Spark's Structured Streaming library that transforms existing Spark SQL computations into continuously running streaming queries. Structured Streaming supports a large majority of Spark primitives including maps, filters, aggregations, and joins. To convert a batch query to a streaming query, users only need to change a single line of dataset reading/writing code and can keep all other computational logic in place. We extend this easy to use API to web serving by creating novel paired sources and sinks that manage a service layer. Intuitively speaking, a web service is a streaming pipeline where the data source and the data sink are the same HTTP request.

Spark Serving can deploy any Spark computation as a web service including all of our contributions (CNTK, LightGBM, SparkML, Cognitive Services, HTTP Services), arbitrary Python, R, Scala, Java, and all compositions and combinations therein. Through other open source contributions in the Spark ecosystem, frameworks such as Tensorflow, XGBoost, and Scikit-learn models join this list.

\subsubsection{Implementation and Architecture}

Under the hood, each Spark worker/executor manages a web service that en-queues incoming data in an efficient parallel data structure that supports constant time routing, addition, deletion, and load balancing across the multiple threads. Each worker node manages a public service for accepting incoming requests, and an internal routing service to send response data to the originating request (on a potentially different node after a shuffle). The worker services communicate their locations and statuses to a monitor service on the driver node. This lets future developers create hooks for their own load balancers, and lets users understand the state and locations of their servers.

If one uses an external load balancer, each request is routed directly to a worker, skipping the costly hop from head node to worker node that is common in other frameworks. The worker converts each request to our Spark SQL type for an HTTP Request (The same types used in HTTP on Spark), and a unique routing address to ensure the reply can route to the originating request. Once the data is converted to Spark's SQL representation it flows through the computational pipeline like any other Spark data.

\begin{figure}[ht]
\vskip 0.1in
\begin{center}
\centerline{
\includegraphics[width=\columnwidth]{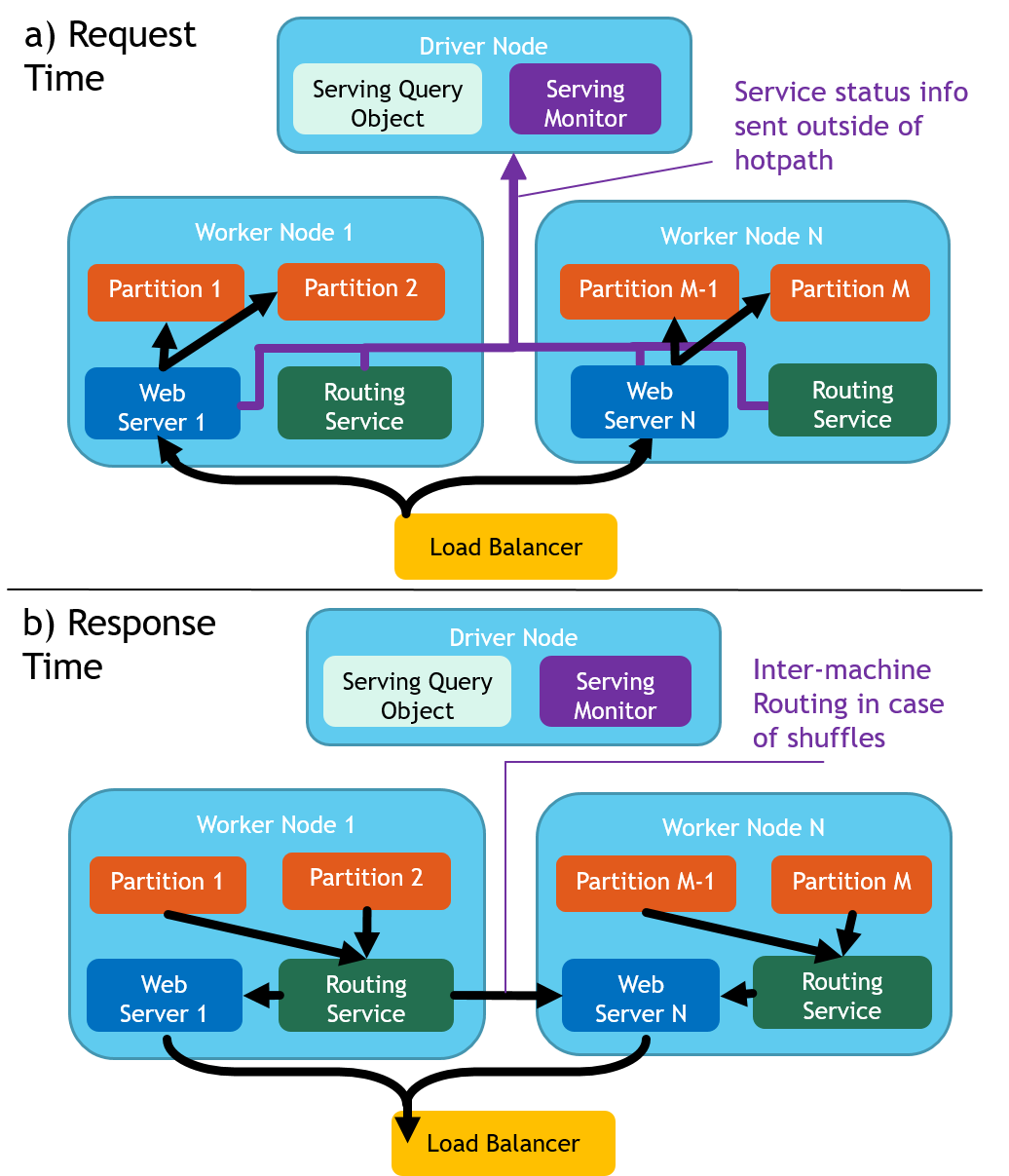}
}
\caption{Overview of Spark Serving architecture during request time (a) and response time (b). Requests are routed to each worker through a load balancer where they can flow through the pipeline when partitions are formed. At reply time, workers use paired ID rows to route information to the correct machines that can reply to the open request. If the request resides on the same machine, Spark Serving uses function dispatch instead of inter-process communication}
\label{fig:serving-architecture}
\end{center}
\vskip -0.3in
\end{figure}

To reply to incoming requests, Spark Serving leverages a new data sink that uses routing IDs and SQL objects for HTTP Responses to reply to the stored request. If the response ends up on a different machine then the originating request, a routing service sends the request to the appropriate machine through an internal network. If the request and response are on the same machine, Spark Serving uses faster function dispatch to reply. In the future, we hope to explore whether the routing service could leverage Spark's underlying shuffle capabilities to speed data transfer throughput. Our choice of HTTP Request and Response Data types enables users to work with the entire breadth of the HTTP protocol for full generality and customizability. Figure \ref{fig:serving-architecture} depicts a schematic overview of the architecture of Spark Serving, and the hotpath during request and response time.

As of MMLSpark v0.14, we have integrated Spark Serving with Spark's new Continuous Processing feature. Continuous Processing dramatically reduces the latency of streaming pipelines from $\approx100ms$ to $ ~ 1ms$. This acceleration enables real-time web services and machine learning applications. Minibatch processing can still be used to maximize throughput, and MMLSpark also provides a batching API for use in continuous processing. 

To the authors' knowledge, Spark Serving is the only serving framework that leverages an existing Spark cluster's workers to serve models. As a result, developers do not need to re-implement and export their models into other languages, such as MLeap, to create web services \cite{mleap}. Furthermore, frameworks that require additional run-times add complexity and incur costs from additional deployment systems. Other serving frameworks, such as Clipper and Azure Machine Learning Services (AML), rely on using Spark's batch processing API to evaluate data. This approach results in large latencies, as each call re-builds the computation graph on the head node, generates code for the workers on the head node, sends this generated code (and all data in the closures) to the workers, sends all the request data to the workers, and collects the response data from the workers. Because we leverage Structured Streaming and continuous processing, all code generation and graph building occurs only in initialization, and the hotpath of the request stays clear of unnecessary computations. Table \ref{table:serving-latencies} demonstrates our improvements in latency over other systems in the literature. Furthermore Spark Serving inherits desire-able fault tolerance behavior, management APIs, and ability to handle shuffles and joins from Spark Streaming. This flexibility makes Spark Serving significantly more general than other frameworks whose workers cannot communicate with each other and do not have access to continuously updated distributed tables. Table \ref{table:serving-features} shows a comparison of functionality across different serving frameworks.

\begin{table}[t]
\caption{Latencies of various Spark deployment methods in milliseconds. We compare Azure Machine Learning Services (AML), Clipper, and MLeap against Spark Serving with local services on an azure virtual machine for reliability. We explore the latency with a pipeline of string indexing, one hot encoding and logistic regression (LR), and a SQLTransformer select statement (SQL). Note that MLeap cannot export SQLTransformer SparkML models. }
\label{table:serving-latencies}
\vskip 0.15in
\begin{center}
\begin{small}
\begin{sc}
\begin{tabular}{lll}
\toprule
Method & LR & Select \\
\midrule
AML           & 530.3 $\pm$ 32.1 & 179.5 $\pm$ 19.5 \\
Clipper       & 626.8 $\pm$ 332.1 & 403.6 $\pm$ 280.0\\ 
MLeap         & 3.18 $\pm$ 1.84 & n/a\\
Spark Serving & \textbf{2.13 $\pm$ .96}  &  \textbf{1.81 $\pm$ .73} \\
\bottomrule
\end{tabular}
\end{sc}
\end{small}
\end{center}
\vskip -0.2in
\end{table}

\begin{table}[t]
\caption{Comparison of features across different Spark Deployment systems}
\label{table:serving-features}
\vskip 0.1in
\begin{center}
\begin{sc}
\begin{tabularx}{\columnwidth}{p{1.7cm}cccp{1.5cm}}
\toprule
Feature     & AML     & Clipper & MLeap & Spark Serving \\
\midrule
Any SparkML & $\surd$ &  $\surd$ & $\times$ & $\surd$ \\
\hline
Any Spark (Narrow)   & $\surd$ &  $\surd$ & $\times$ & $\surd$ \\ 
\hline
Joins       & $\times$ &  $\times$ & $\times$ & $\surd$ \\ 
\hline
Aggregates  & $\times$ &  $\times$ & $\times$ & $\surd$ \\
\hline
Shuffles    & $\times$ &  $\times$ & $\times$ & $\surd$ \\
\hline
Millisecond Latency    & $\times$ &  $\times$ & $\surd$ & $\surd$ \\

\bottomrule
\end{tabularx}
\end{sc}
\end{center}
\vskip -0.25in
\end{table}

\section{Applications}
\label{Applications}

We have used MMLSpark to power engagements in a wide variety of machine learning domains, such as text, image, and speech domains. In this section, we highlight the aforementioned contributions in our ongoing work to use MMLSpark for wildlife conservation and custom search engine creation.

\subsection{Snow Leopard Conservation}

Snow leopards are dwindling due to poaching, mining, and retribution killing yet we know little about how to best protect them. Currently, researchers estimate that there are only about four thousand to seven thousand individual animals within a potential 2 million square kilometer range of rugged central Asian mountains \cite{slt}. Our collaborators, the Snow Leopard Trust, have deployed hundreds of motion sensitive cameras across large areas of Snow Leopard territory to help monitor this population \cite{slt}. Over the years, these cameras have produced over 1 million images, but most of these images are of goats, sheep, foxes, or other moving objects and manual sorting takes thousands of hours. Using tools from the MMLSpark ecosystem, we have created an automated system to classify and localize Snow Leopards in camera trap images without human labeled data. This method saves the Trust hours of labor and provides data to help identify individual leopards by their spot patterns.

\begin{figure}[ht]
\vskip 0.05in
\begin{center}
\centerline{
\includegraphics[width=\columnwidth]{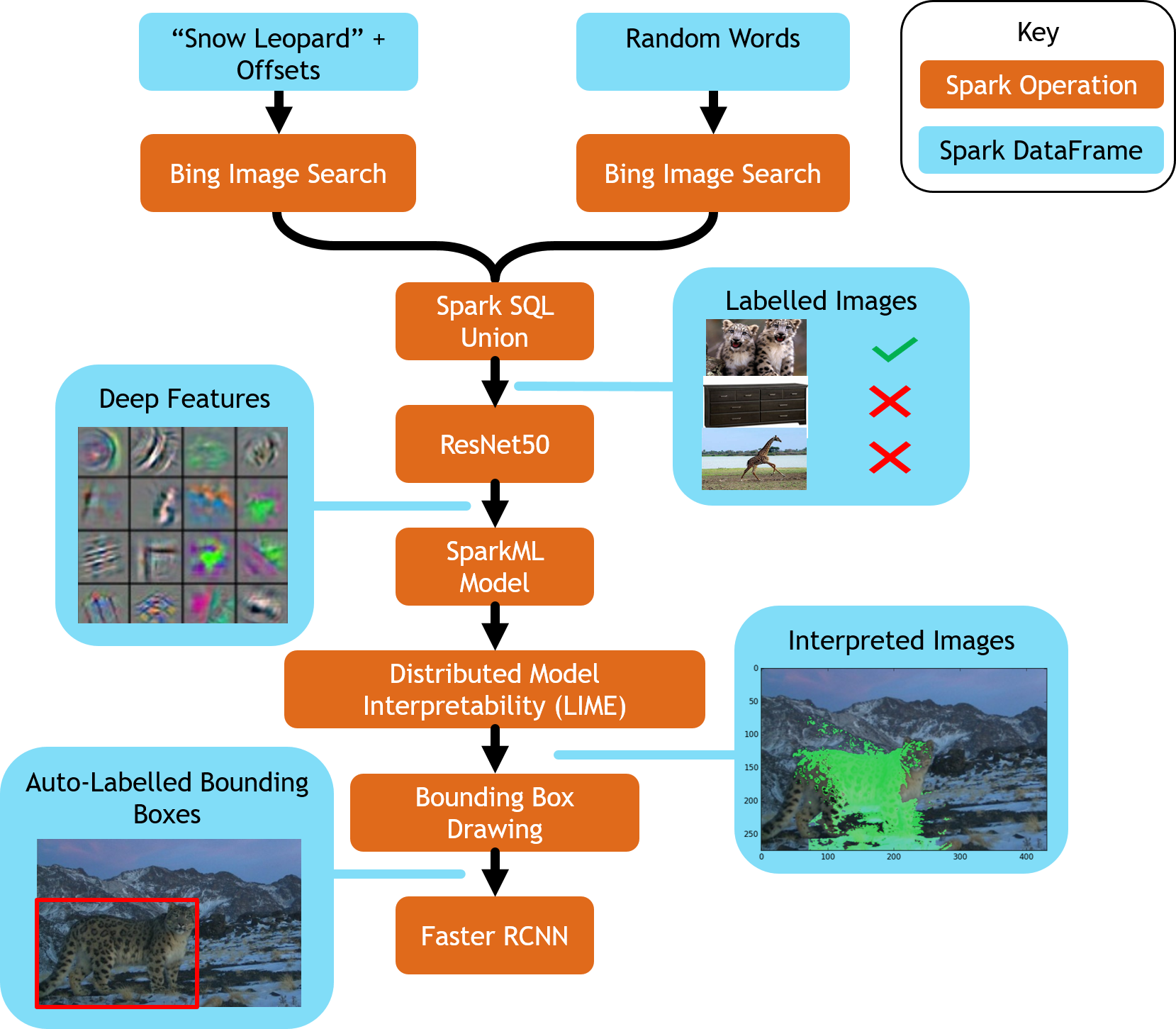}
}
\caption{End to end architecture of unsupervised object detection on Spark}
\label{fig:architecture}   
\end{center}
\vskip -0.2in
\end{figure}

\begin{figure}[ht]
\vskip 0.2in
\begin{center}
\centerline{
\includegraphics[width=\columnwidth]{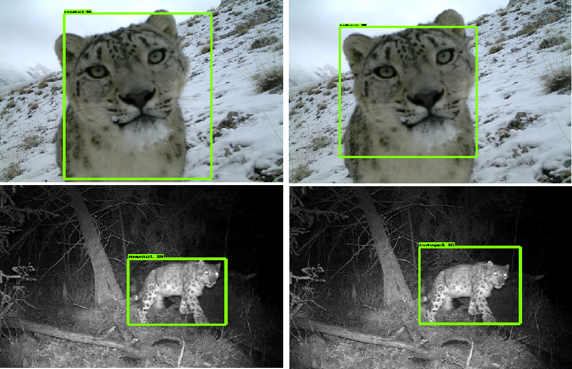}
}
\centerline{
\includegraphics[width=\columnwidth]{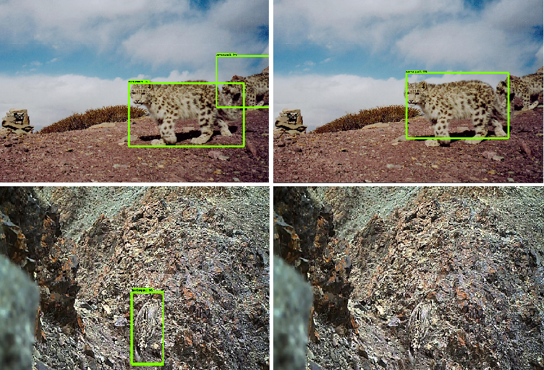}
}
\caption{Human labeled images (top left) and unsupervised Faster-RCNN outputs (top right). Difficult human labeled images (bottom left) and unsupervised Faster-RCNN outputs (bottom right).}
\label{fig:bboxcomp}
\end{center}
\vskip -0.3in
\end{figure}

\subsubsection{Unsupervised Classification}
\label{unsupervised-classification}
In our previous work, we used deep transfer learning with CNTK on Spark to create a system that could classify Snow Leopards in Camera trap images \cite{mmlspark}. This work leveraged a large dataset of manually labeled images accumulated through years of intensive labelling by the Snow Leopard Trust. In this work, we show that we can avoid all dependence on human labels by using Bing Image Search to automatically curate a labeled Snow Leopard dataset. More specifically, we used our SparkML bindings for Bing Image Search to make this process easy and scalable. To create a binary classification dataset, we first create a dataset of leopards by pulling the first 80 pages of the results for the ``Snow Leopard'' query. To create a dataset of negative images, we drew inspiration from Noise Contrastive Estimation, a mathematical technique used frequently in the Word Embedding literature \cite{noise-contrastive}. More specifically, we generated a large and diverse dataset of random images, by using random queries as a surrogate for random image sampling. We used an existing online random word generator to create a dataframe of thousands of random queries. We used Bing Images on Spark to pull the first 10 images for each random query. After generating two datasets, we used Spark SQL to add labels, stitch them together, drop duplicates, and download the images to the cluster in only a few minutes. Next, we used CNTK on Spark to train a deep classification network using transfer learning on our automatically generated dataset. Though we illustrated this process with Snow Leopard classification, the method applies to any domain indexed by Bing Images.

\subsubsection{Unsupervised Object Detection}

 Many animal identification systems, such as HotSpotter, require more than just classification probabilities to identify individual animals by their patterns \cite{hotspotter}. In this work, we introduce a refinement method capable of extracting a deep object detector from any image classifier. When combined with our unsupervised dataset generation technique in Section \ref{unsupervised-classification}, we can create a custom object detector for any object found on Bing Image Search. This method leverages our LIME on Spark contribution to ``interpret'' our trained leopard classifier. These classifier interpretations often directly highlight leopard pixels, allowing us to refine our input dataset with localization information. However, this refinement operation incurs the 1000x computation cost associated with LIME, making even the distributed version untenable for real-time applications. However, we can use this localized dataset to train a real-time object detection algorithm like Faster-RCNN \cite{fasterrcnn}. We train Faster-RCNN to quickly reproduce the computationally expensive LIME outputs, and this network serves as a fast leopard localization algorithm that does not require human labels at any step of the training process. Because LIME is a model agnostic interpretation engine, this refinement technique can apply to any image classification from any domain. Figure \ref{fig:architecture} shows a diagram of the end-to-end architecture.
 
\begin{table}[t]
\caption{Object detector performance (mAP@[.5:.95]) evaluated on human curated test images}
  \label{table:result-detector}
\vskip 0.15in
\begin{center}
\begin{small}
\begin{sc}
\begin{tabular}{lll}
\toprule
Training method & mAP   \\
\midrule
Unsupervised + Pre-training & 49.8 \\
Human     & 30.9 \\
Human + Pre-training    & 79.3       \\
\bottomrule
\end{tabular}
\end{sc}
\end{small}
\end{center}
\vskip -0.3in
\end{table}

\begin{table}[t]
\caption{Deep classifier performance on synthetic data, human labelled data, and a logistic regression model on human data}
\label{table:result-classifier}
\vskip 0.15in
\begin{center}
\begin{small}
\begin{sc}
\begin{tabular}{lll}
    \toprule
    Algorithm & Accuracy   \\
    \midrule
    Unsupervised & 77.6\% \\
    Human        & 86.8\% \\
    Human + LR & 65.6\%       \\
    \bottomrule
\end{tabular}
\end{sc}
\end{small}
\end{center}
\vskip -0.3in
\end{table}

\subsubsection{Results for Unsupervised Leopard Detection}

We discovered that our completely unsupervised object detector closely matched human drawn bounding boxes on most images. Table \ref{table:result-detector} and \ref{table:result-classifier} show that our method can approach that of a classifiers and object detectors trained on human labelled images. However, certain types of images posed problems for our method. Our network tended to only highlight the visually dominant leopards in images with more than one leopard, such as those in Figure \ref{fig:bboxcomp}. We hypothesize that this arises from our simple method of converting LIME outputs to bounding boxes. Because we only draw a single box around highlighted pixels, our algorithm has only seen examples with a single bounding box. In the future, we plan to cluster LIME pixels to identify images with bi-modal interpretations. Furthermore, the method also missed several camouflaged leopards, as in Figure \ref{fig:bboxcomp}. We hypothesize that this is an anthropic effect, as Bing only returns clear images of leopards. We plan to explore this effect by combining this Bing generated data with active learning on a ``real'' dataset to help humans target the toughest examples quickly. 

\begin{figure}[ht]
\vskip 0.1in
\begin{center}
\centerline{
\includegraphics[width=\columnwidth]{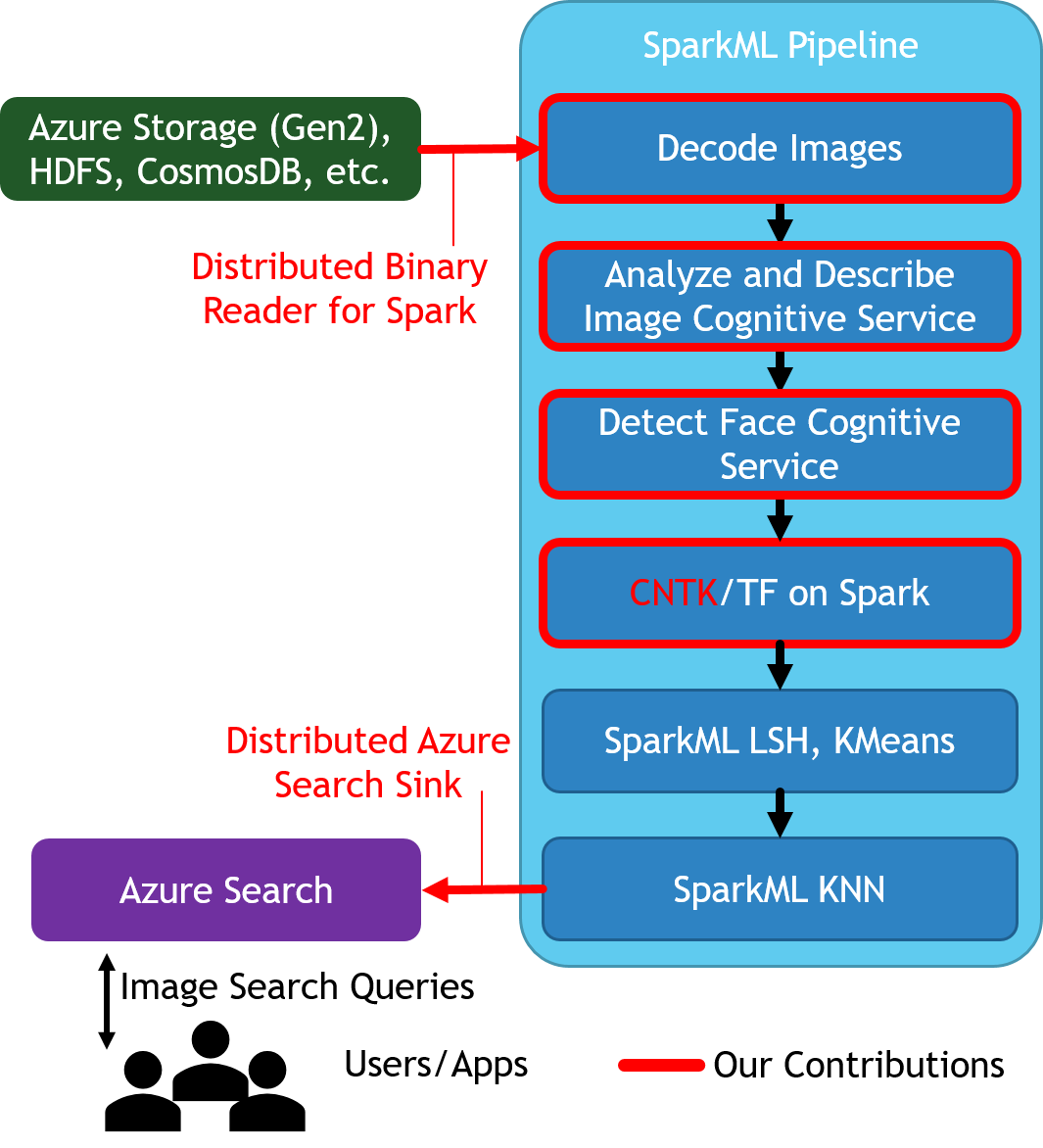}
}
\caption{Overview of distributed image analysis pipeline that leverages containerized microservices on Spark, deep-learning computation graphs, nontrivial joins for KNN and locally sensitive hashing, and distributed index creation with Azure Search.}
\label{fig:search-architecture}
\end{center}
\vskip -0.4in
\end{figure}

\subsection{Visual Search Engine Creation}

When many different frameworks unify within the same API and ecosystem, it becomes possible to create high-quality distributed applications with very few lines of code. We demonstrate this by using MMLSpark and its surrounding ecosystem to create a visual search engine. As is shown in Figure \ref{fig:search-architecture}, we use MMLSpark's binary reading extension to ingest raw files with high throughput using all nodes of the cluster. We can then pump these images through a variety of Computer Vision services with MMLSpark's cognitive service integration. These services add image content, descriptions, faces, celebrities, and other useful intelligence to the dataframe of images. We can then featurize images with headless networks such as ResNet50 or InceptionV3 with either CNTK on Spark (ours), or Tensorflow on Spark (Databrick's). We can then pass these high-level features through SparkML's locally sensitive hashing implementation, K-means clustering, or a third-party K-nearest neighbor SparkML package \cite{spark-knn}. We can then create and write the dataframe to an Azure Search index in a distributed fashion in a single line with MMLSpark's Azure Search integration. The resulting index can be quickly queried for fuzzy matches on image information, content, and visually similarity to other images. 

\section{Conclusion}
\label{Conclusion}

In this work we have introduced Microsoft Machine Learning for Apache Spark, a framework that integrates a wide variety of computing technologies into a single distributed API. We have contributed CNTK, LightGBM, and LIME on Spark, and have added a foundational integration between Spark and the HTTP Protocol. We built on this to integrate the Microsoft Cognitive Services with Spark and create a novel real-time serving framework for Spark models. We have also shown that through combining these technologies, one can quickly create and deploy intelligent applications. Our first application used Bing, CNTK, LIME, and Spark Serving to create a deep Snow Leopard detector that did not rely on costly human labeled data. This application made no assumptions regarding the application domain and could extract custom object detectors for anything searchable on Bing Images. Our second application leveraged containerized cognitive services, HTTP on Spark, CNTK, Tensorflow, and other tools from the SparkML ecosystem to create an intelligent image search engine. The contributions we have provided in MMLSpark allow users to draw from and combine a wide variety machine learning frameworks in only a few lines of Spark code. This work dramatically expands the Spark framework into several new areas of modern computing and has helped us rapidly create several distributed machine learning applications.

\section{Future Work}

We hope to continue adding computation frameworks and aim to extend the techniques of HTTP on Spark to other protocols like GRPC. We are continuously expanding our collection of HTTP-based Spark models, and aim to add services from Google, IBM, and AWS in addition to Microsoft. We also hope to explore Spark Clusters of accelerated hardware SKUs like cloud TPUs and FPGAs to accelerate computations. Additionally, we aim to leverage advancements in the Spark Ecosystem such as Barrier Execution to improve the fault tolerance of LightGBM \cite{barrier-execution}. Additionally, we hope to integrate Spark Serving as a deployment flavor for the popular machine learning  management tool, MLFlow \cite{mlflow}. 

% Acknowledgements should only appear in the accepted version.
\section*{Acknowledgements}
We would like to acknowledge the generous support from our collaborators, Dr. Koustubh Sharma, Rhetick Sengupta, Michael Despines, and the rest of the Snow Leopard Trust. We would also like to acknowledge those at Microsoft who helped fund and share this work: the Microsoft AI for Earth Program, Lucas Joppa, Joseph Sirosh, Pablo Castro, Brian Smith, Arvind Krishnaa Jagannathan, and Wee Hyong Tok.

\bibliography{example_paper}
\bibliographystyle{icml2019}

\end{document}